\documentclass[3p,a4paper]{elsarticle}
\usepackage[english]{babel}
\usepackage[utf8]{inputenc}
\usepackage[cmex10]{amsmath}
\usepackage{xcolor}
\usepackage{graphicx}
\usepackage{array}
\usepackage{mdwmath}
\usepackage{mdwtab}
\usepackage{eqparbox}
\usepackage{pdflscape}
\usepackage{pbox}
\usepackage{color}
\usepackage{nccmath}
\usepackage{amssymb}
\usepackage{multirow}
\usepackage{longtable}
\usepackage{array,multirow,graphicx}
\usepackage{subfig}
\usepackage{lscape}
\usepackage{setspace}
\usepackage{todonotes}
\usepackage{longtable}
\journal{Neurocomputing}
\usepackage{url}

\begin{document}

\begin{frontmatter}

\title{Introducing Randomized High Order Fuzzy Cognitive Maps as Reservoir Computing Models: A Case Study in Solar Energy and Load Forecasting}

\author[minds]{Omid Orang\corref{mycorrespondingauthor}}
   \cortext[mycorrespondingauthor]{Corresponding author}
   \ead{omid.orang2009@gmail.com}
\author[ifnmg,minds]{Petr\^onio C\^andido de Lima e Silva}
   \ead{petronio.candido@ifnmg.edu.br}
\author[dee,minds]{Frederico Gadelha Guimar\~aes}
   \ead{fredericoguimaraes@ufmg.br}
   \ead[url]{https://minds.eng.ufmg.br/}
 
\address[minds]{Machine Intelligence and Data Science (MINDS) Laboratory, Federal University of Minas Gerais, Belo Horizonte, Brazil}
\address[ifnmg]{Federal Institute of Education Science and Technology of Northern Minas Gerais, Janu\'aria Campus, Brazil}
\address[dee]{Department of Electrical Engineering, Universidade Federal de Minas Gerais, Belo Horizonte, Brazil}

\begin{abstract}
\begin{singlespace}
\noindent 
Fuzzy Cognitive Maps (FCMs) have emerged as an interpretable signed weighted digraph method consisting of nodes (concepts) and weights which represent the dependencies among the concepts. Although FCMs have attained considerable achievements in various time series prediction applications, designing an FCM model with time-efficient training method is still an open challenge. Thus, this paper introduces a novel univariate time series forecasting technique, which is composed of a group of randomized high order FCM models labeled R-HFCM. The novelty of the proposed R-HFCM model is relevant to merging the concepts of FCM and  Echo State Network (ESN) as an efficient and particular family of Reservoir Computing (RC) models, where the least squares algorithm is applied to train the model. From another perspective, the structure of R-HFCM consists of the input layer, reservoir layer, and output layer in which only the output layer is trainable while the weights of each sub-reservoir components are selected randomly and keep constant during the training process. As case studies, this model considers solar energy forecasting with public data for Brazilian solar stations as well as Malaysia dataset, which includes hourly electric load and temperature data of the power supply company of the city of Johor in Malaysia. The experiment also includes the effect of the map size, activation function, the presence of bias and the size of the reservoir on the accuracy of R-HFCM method. The obtained results confirm the outperformance of the proposed R-HFCM model in comparison to the other methods. This study provides evidence that FCM can be a new way to implement a reservoir of dynamics in time series modelling.
\end{singlespace}

\end{abstract}

\begin{keyword}
Time series forecasting \sep Fuzzy Cognitive Maps \sep Reservoir Computing \sep Echo State Network \sep Least Squares algorithm
\end{keyword}

\end{frontmatter}


\section{Introduction}
\label{sec:introduction}

In the past decades, time series analyzing and forecasting, as a hot research topic, has attracted the attention of many researchers to make better decisions across various disciplines specially with the advent of some recent technologies such as loT, IoE or Big Data. Despite the proposal of plenty of various forecasting procedures in the literature, the accurate prediction of the future values is still an open challenge due to the existence of non-linearity and uncertainty in real time events. Statistical time series forecasting methods are limited due to some characteristics such as time consuming, lack of scalability and explainability, inability to deal with uncertainty and complex problems in the real world \cite{Silva2019Thesis}. Accordingly, fuzzy time series (FTS) was presented by Song and Chissom in \cite{song1993forecasting,Song1994,Song1997} to handle the mentioned problems. Numerous FTS forecasting models have been introduced in the literature in many applications due to some important features including simplicity, explainability, flexibility, updatability, readability, versatility, and high accuracy of forecasting \cite{Silva2019Thesis,bose2019designing}. 

Fuzzy Cognitive Map (FCM) was proposed by Kosko in \cite{Kosko1986} as weighted knowledge-based model that is used to do the task of forecasting by extracting knowledge in FTS models. FCM is a kind of interpretable recurrent neural network that combines the concepts of fuzzy logic and neural networks which is composed of nodes (representing concepts) and signed directed relations (causal relations) between a couple of concepts. Noteworthy that FCMs as qualitative soft computing techniques can be used to represent the dynamic behaviour of complex systems with high ability to dealing with uncertainties \cite{vanVliet2010}. Therefore a wide range of FCM-based time series forecasting methods have been developed in the literature \cite{Stach2008both,Wojciech2009comprative,PAPAGEORGIOU201228,lu2013linguistic,Homenda2016clusteringFCM,Shanchao2018WHFCM,Papageorgiou2019FCMNNGAS,vanhoenshoven2018fuzzy,vanhoenshoven2020pseudoinverse,yuan2020timekernelHFCM}.

In general, time series prediction by FCM utilization consists of two stages \cite{gao2020robustemwt}. Firstly, designing the appropriate structure of FCM using common strategies including granularity \cite{Stach2008both}, membership values representation \cite{song2010fuzyyNNFCM} and Fuzzy c-means clustering \cite{Lu2014HFCMCMEANS}. Also, the authors in \cite{LIU2020106105EMDHFCM,Shanchao2018WHFCM} employed wavelet transformation and empirical mode decomposition (EMD) to formulate the structure of FCM. Secondly, learning weight matrices. As the literature reviewed \cite{salmeron2019learning, felix2019review}, population-based methods play a major role in FCM learning methods compared with Hebbian-based, hybrid and other methods. Genetic Algorithm (GA) \cite{Froelich2016GranularFCM}, Real coded genetic algorithm (RCGA) in \cite{lu2013linguistic}, Particle Swarm Optimization in \cite{Homenda2014}, Simulated Annealing (SA) in \cite{ghazanfari2007comparing}, Game-based learning model in \cite{Lue2009gamebased}, Immune Algorithm (IA) in \cite{lin2009immune}, Big Bang-Big Crunch (BB-BC) in \cite{Yesli2010bigbang}, Ant Colony Optimization (ACO) \citep{ding2011first},  Artificial Bee Colony (ABC) algorithm in \cite{Yesil2013FuzzyCM}, Cultural Algorithm (CA) in \cite{Ahmadi2014cultural}, Imperialist Competitive Learning Algorithm (ICLA) in \cite{Ahmadi2015Imperialistcognitive} have all been used as robust and accurate FCM learning methods. In addition to these, other strategies have been used as well, such as Multi-objective optimization algorithm so-called MOEA-FCM in \cite{chi2015MOEA}, dynamic multi-agent genetic algorithm (DMAGA) proposed in \cite{liu2015dynamic}, evolutionary multi-tasking multi-objective memetic FCMs (MMMA-FCMs) learning algorithm adopted in \cite{Shen2019EvolutionaryMFCM}, Inactivation-based batch many-task  evolutionary algorithm (IBMTEA-FCM) in \cite{articleWang2021manytask} are some examples of population-based learning methods. 

Based on the literature, GA and PSO  have been used widely to learn weight matrices in FCM-based time series forecasting models using evolutionary-based learning methods. For example, \cite{Homenda2014} proposed a univariate forecasting method using FCM and C-means clustering plus moving window technique, while PSO algorithm was used to learn weight matrices. In \cite{lu2013linguistic}, the authors proposed univariate time series forecasting using FCM and C-means clustering in which real coded genetic algorithm (RCGA) was employed to extract weight matrix. \cite{Lu2014HFCMCMEANS} introduces a time series forecasting model based on the synergy of High-Order FCM (HFCM) and fuzzy C-means clustering in which PSO is applied to optimize the weights of HFCM. In \cite{Stach2008both} a two-level forecasting technique was proposed to carry out forecasting in both numerical and linguistic terms using RCGA learning method to train weight matrices. A double-phased approach was introduced in \cite{Froelich2016GranularFCM} combining granular FCM and fuzzy C-means clustering which is trained through GA algorithm. The authors in \cite{Orang2020} proposed univariate HFCM-FTS model using GA to train weight matrices. Intuitionistic fuzzy grey cognitive maps for forecasting interval-valued time series learned via Differential Evolution (DE) was proposed in \cite{hajek2020intuitionistic}. 

Since the population-based learning methods are much time consuming and computationally expensive due to the large number of learning parameters and large number of learning processes, finding and designing fast and robust learning strategies is considered the main challenges in this area. Therefore, evolutionary learning has been replaced with other techniques in some references. For instance, ridge regression in \cite{Shanchao2018WHFCM,yuan2020timekernelHFCM,wu2019time}, Bayesian ridge regression in \cite{LIU2020106105EMDHFCM}, Moore-Penrose inverse in \cite{vanhoenshoven2020pseudoinverse}. A rapid and robust learning method with maximum entropy was proposed in \cite{Feng2021TheLOentropy} to learn large scale FCMs composed of least-squares and maximum entropy terms. The robustness of the well-learned FCM is guaranteed by Least-squares term and the maximum entropy term regularizes the distribution of the weights of the well-learned FCM. Recently, a new time series modeling based on least square FCM termed as LSFCM was introduced in \cite{a14030069}. In this method Fuzzy c-means clustering is exploited to construct FCM concepts and the weight matrices are adopted from the given historical observation of time series using the least square method which is much faster than other population-based methods. 

The crux objective of our research study is to develop a novel univariate FCM-based time series forecasting model focusing mainly on FCM training to promote the accuracy and efficacy of time series methods applying FCM. Thus, the concepts of High Order FCM (HFCM) and Echo State Network (ESN) reservoir computing are merged to construct a novel forecasting technique using the least square training algorithm termed as Randomized HFCM (R-HFCM). More vividly, this model consists of three layers: input layer, reservoir and output layer. The reservoir layer consists of a group of HFCM-FTS models proposed in \cite{Orang2020} but in this case the weights are randomly initialized such that the Echo State Property (ESP) condition in ESN reservoir computing is satisfied. Then, each sub-reservoir generates its output independently after feeding input into each sub-reservoir separately. Finally, the least squares minimization technique is applied to train the output layer and generate the final predicted value. As the computational experiments reveal, in addition to improving the accuracy, the proposed model is much faster than HFCM-FTS method. Furthermore, the results are promising in comparison to other state-of-the-art techniques in the literature. 

The remainder of this paper is organized as follows: Section \ref{sec:Preliminaries} presents  a brief description of  the  Fuzzy Cognitive Map (FCM) and Echo State Network (ESN); Section \ref{sec:GRHFCM} introduces the proposed method in details; Section \ref{sec:experiments} presents the experimental results and discussion and finally the paper conclusion and some possibilities of future works are drawn in Section \ref{sec:conclusion}.

\section{Preliminaries}
\label{sec:Preliminaries}

\subsection{Fuzzy Cognitive Maps}
Fuzzy Cognitive Map (FCM) was proposed by \cite{Kosko1986} as qualitative and causal model which combines fuzzy logic and CM to present uncertainties and complex characteristics of the systems. Graphically speaking, FCM is powerful interpretable knowledge-based model which is composed of concepts (nodes) and the causal connection among the pair of concepts  . The directed signed arrows among the concepts is known as weights which reflect the effect of one node on another one. Therefore, FCM is based on cause and effect causality. If the node influences on others is cause and while influenced by others is an effect.

\begin{figure}[!htb]
    \centering
    \includegraphics[width=0.7\textwidth]{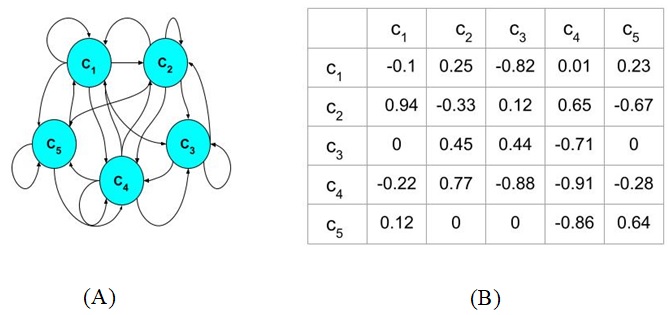}
    \caption{Simple FCM with 5 nodes (A) graphical structure (B) Weight matrix}
    \label{fig:FCMrepresentation}
\end{figure}

Each FCM is identified by four elements which is defined by 4-tuple $(\mathbf{C}, \mathbf{W}, \mathbf{a}, f)$, where $\mathbf{C} = [c_1 ,\ldots,c_n ]$ is the set of $n$ concepts, which are the variables (and the nodes of the graph) that compose the system. The state values (activation degree) of these concepts at any time $t$ is represented as follows:

\begin{equation}\label{eq:a}
\mathbf{a}=(a_1, \ldots, a_n )
\end{equation}
where $a_i \in [0,1]$, $i=1,2,...,n$. The relations between the nodes are described as an $n\times n$ weight matrices as follows:

\begin{equation}\label{eq:W}
\mathbf{W}=\left(
\begin{array}{ccc}
w_{11}&\ldots& w_{1n}\\
\vdots&\ddots &\vdots\\
w_{n1}&\ldots& w_{nn}
\end{array}\right)
\end{equation}
where $w_{ij} \in [-1,1]$ ($i,j=1,2,\dots,n$) indicates the relationship between the source and the target nodes which can be positive, negative or zero. If $w_{ij}>0$, a decrease/increase in value of node $c_i $ makes a decrease/increase in value of node $c_j$. If $w_{ij}<0$, a decrease/increase in value of node $c_i $ makes an increase/decrease in value of node $c_j$. If $w_{ij}=0$, there exists no relation between nodes $c_i$ and $c_j$. Figure \ref{fig:FCMrepresentation} exhibits the simple example of FCM structure including 5 nodes. 

The final element is activation function ($f$) used in Kosko's updating rule to preserve the activation degree of each concept inside the predefined interval. Bivalent, trivalent, hyperbolic tangent and the sigmoid are the most common activation functions in the literature \cite{felix2019review}. According to the Kosko's activation rule, the activation degree of each node at $t+1$-th iteration is updated considering the weight matrix and activation degree of all connected nodes at the $t$-th iteration. In other words, the weights are used to transfer the current activation state of each concept to the next. Hence, the following formula describes the dynamics of FCM.

\begin{equation}\label{eq:activation_rule}
a_i(t+1)=f\left( \sum_{j=1}^{n} w_{ij} a_j(t) \right)
\end{equation}
where $w_{ij} $ highlights the value of the causal relationship between concepts $ c_i$ and $ c_j$, whereas $a_i^t$ represents the state value of concept $c_i$ at time step $t$.

Although the aforementioned updating rule has been utilized widely in many FCM-based applications, other modifications have been proposed in the literature with/without self-connection and with/without considering  memory \cite{papakostas2010classifying,felix2019review,Stylios2004complex,Parsopoulos2002pso,Papageorgiou2011ruleextraction}. 

Dynamically speaking, the equation \eqref{eq:activation_rule} expresses only the first order dynamics of the FCM. It means that the value of each concept at time $t+1$ only relies on the activation level of all concepts at time $t$. But an accurate modeling of FCM will not be obtained only by considering the current activation value of the concepts and ignoring the past values. Accordingly, High-Order FCMs (HFCMs) were proposed to improve the performance of FCMs and cover the mentioned limitation. Thus, the equation \eqref{eq:activation_rule} is modified to describe the dynamic behavior of complex systems more accurately. The below equation describes the $k$-order FCM: \cite{Shanchao2018WHFCM,Lu2014HFCMCMEANS}

\begin{equation}\label{eq:activation_rule_2}
a_i(t+1)=f\left(w^0_{i} + \sum_{j=1}^{n} w^1_{ij} a_j(t)+w^2_{ij} a_j(t-1)+ \ldots +w^{\Omega}_{ij} a_j(t-\Omega+1) \right)
\end{equation}
where $w^0_{i}$ stands for the bias term and $w^{\Omega}_{ij}$ is the casual relation originating from  $c_i$ and pointing to $c_j$ at time step $t-\Omega+1$. Based on this equation, the activation level of $i$-th node at the moment $t+1$ depends on the activation degree of all concepts at $\{t,t-1,..,t-\Omega+1\}$ moments, not only the activation states of the concepts at time $t$, during the iterative process.

Equation \eqref{eq:activation_rule_2} can be rewritten in matrix form as:
\begin{equation}
\mathbf{a}(t+1) = f\left( \mathbf{w}^0 + \mathbf{W}^1\mathbf{a}(t) + \mathbf{W}^2 \mathbf{a}(t-1) + \ldots + \mathbf{W}^{\Omega} \mathbf{a}(t-\Omega+1)  \right)
\end{equation}

\subsection{Echo State Network (ESN)}

Echo state network (ESN) was proposed by Jaeger \cite{jaeger2001echo} as one of the improved Recurrent Neural Network (RNN).  Standard RNN has been considered as a powerful tool to simulate complex dynamic systems, but training algorithms of RNNs involve some downsides including: relatively high computational training costs and potentially slow convergence, local minima of the error function and vanishing of the gradients. With the goal of handling these issues, a new approach in RNN training called Reservoir Computing (RC) was proposed to study of initialization conditions and stability instead of focusing on the training algorithm. ESN as one of the popular RC techniques, has a very simple and fast training process and it has abundant nonlinear
echo states and short-term memory because of  high-dimensional projection and highly sparse connectivity of neurons in the reservoir, which are useful for modeling dynamical systems \cite{ma2017deep}.

\begin{figure}[!htb]
    \centering
    \includegraphics[width=0.7\textwidth]{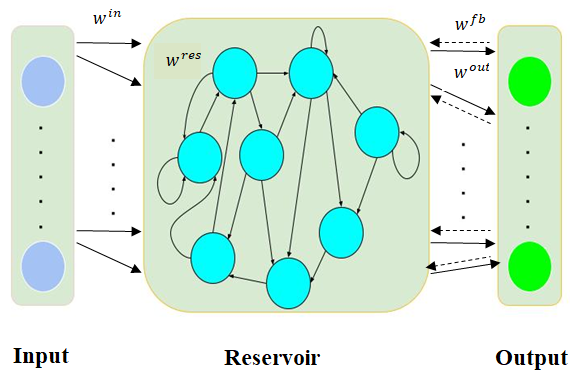}
    \caption{Structure of Traditional ESN}
    \label{fig:ESN}
\end{figure} 

The structure of ESN consists of three basic components including input layer, a large recurrent hidden layer (called the dynamical reservoir as untrained large, sparsely connected nonlinear layer) with fixed sparse hidden-to-hidden connections and an output layer.  Despite traditional RNNs,  the dynamical reservoir or hidden layer in ESNs is untrainable and only the output weights are trained which is the most excellent feature of ESNs. Figure \ref{fig:ESN} highlights the structure of traditional ESN with $M$ input units, $N$ reservoir units and $L$ output (readout) neurons, while $u(t) = (u_1(t),\dots,u_M(t))^T, x(t) = (x_1(t),\dots,x_N(t))^T$, and $y(t) = (y_1(t),\dots,y_L(t))^T$ determine their activation at each time step. The output of dynamic reservoir neurons and the output of ESN at time step $t+1$ are updated according to the following equations respectively.

\begin{equation}\label{eq:DR_output}
 x(t+1)=f(\mathbf{W^{in}}\cdot u(t + 1) +\mathbf{W^{res}}\cdot x(t) + \mathbf{W^{f_b}} \cdot y(t))
\end{equation}
\begin{equation}\label{eq:ESN_output}
 y(t + 1) = f^{out}({\mathbf{W^{out}}} [x(t+1); ~u(t+1)])
\end{equation}
where the $N\times M$ input weight matrix $\mathbf{W^{in}}$ represents the relation among input units to the reservoir neurons; the internal connection among neurons inside the reservoir layer is represented via $N\times N$ weight matrix $\mathbf{W^{res}}$; the output weight matrix is represented via $\mathbf{{W^{out}}}$ with $L \times (M+ N)$ dimension. Finally, an $N \times L$ weight matrix represents the connection projected back from readout neurons to reservoir units as $\mathbf{W^{fb}}$ or $\mathbf{W^{back}}$. 

Interestingly, both internal and input weights are initialized randomly with no changes during the training and testing process. On the other side, a simple linear regression is applied to train readout layer. Noteworthy that an ESN is trained through a supervised learning method in two stages. At the first level, the $M$ dimensional inputs are mapped into a high dimensional reservoir state to reach the echo states $x$. Then a simple regression method is exploited to train the output weights. 

In addition to the mentioned unique features of ESN, regarding the fixed non trainable reservoir and just learning readout layer, the ESN can also act as a kernel in kernel-based learning technique to catch the input dynamics by application of a random high-dimensional projection method which individuates the ESN from other RNNs \cite{ma2017deep}. In a nutshell, the learning of ESN is comparable with other RNNs because it is simple and fast. Besides, it does not trap into local minimum which enables ESN for modeling dynamics of time series. Due to the simple method and high learning efficiency, ESN has been successfully applied to many fields, such as time series prediction tasks \cite{DEIHIMI2013382,li2012chaotic,LIN20097313}, dynamic pattern classification \cite{jaeger2002adaptive,wang2016effective}, speech recognition \cite{tong2007learning,jaeger2007optimization}, and so on. In other words, a very easy training process, high-dimensional projection as well as highly sparse connectivity of neurons in the reservoir enable ESN to modeling dynamic systems. 

\subsubsection{Initialization and Hyper-parameters}

An ESN is constructed according to the significant hyper-parameters including the reservoir size $N$, the input scale $IS$, the spectral radius $\mathbf{\epsilon}$, and sparsity $\mathbf{\gamma}$. The input scale is employed to initialize the input to hidden weight matrix $\mathbf{W^{in}}$ such that each member of $\mathbf{W^{in}}$ obeys the uniform distribution in $[-IS, IS]$. $\mathbf{\gamma}$ determines the proportion of non-zero elements in ($\mathbf{W^{res}}$) and $\mathbf{\epsilon}$ is the spectral radius of $\mathbf{W^{res}}$, which must be set smaller than 1 \cite{ma2017deep}.

As mentioned, in the training process of ESN, just the readout weight matrices are trainable while the $\mathbf{W^{res}}$ is selected randomly with a uniform distribution symmetric around the zero value before training execution to provide the requirements of echo state condition \cite{jaeger2001echo,ma2017deep}. The initialization of $\mathbf{W^{res}}$ is done through the following procedure to assure that the maximum absolute eigenvalue or spectral radius is less than one. In other words, firstly an internal randomly weight matrix $\mathbf{W^{rand}}$ will be generated. Then rescaled to meet the Echo State Property (ESP) condition as below:

\begin{equation}\label{eq:internal_weights}
\mathbf{W^{res}}=\frac {\Huge{\epsilon} \cdot {\mathbf{W^{rand}}}}{\huge{\rho_{max}}(\mathbf{W^{rand})}}
\end{equation}
where $\rho_{max}(\mathbf{W^{rand})}$ denotes the maximum eigenvalue of matrix $\mathbf{W^{rand}}$ when its elements are generated randomly. To guarantee the stability of ESN, $\mathbf{\epsilon}$ must be set smaller than 1. This is a necessary condition of ESN stability \cite{jaeger2001echo}. The final step is to determine $\mathbf{W^{out}}$ which is unknown and alterable. $\mathbf{W^{out}}$ is calculated through the following learning equation:

\begin{equation}\label{eq:output_weights}
\mathbf{W^{out}}= (X^T X)^{-1} X^T Y
\end{equation}
where $(X^T X)^{-1}$ denotes the inversion of square matrix $X^T X$.

Noteworthy, ESNs with leaky integrator neurons \cite{lukovsevivcius2009reservoir}, $\phi$-ESN in \cite{gallicchio2011architectural}, ESNs with circle reservoir topology \cite{Xue2017leakyechostate} and different types of deep ESNs in \cite{gallicchio2016deep,ma2017deep, sun2017deep,zheng2020long} represent powerful variations and advances in ESN in the literature.

\section{Proposed R-HFCM method}
\label{sec:GRHFCM}

As was mentioned earlier, the large number of processes, time-consuming and learning parameters adjustment are counted as the major deficiencies of the population-based FCM learning methods. Thus, to rectify the above-mentioned limitations, designing a faster training model seems vital. Due to that, some publications focused on other new techniques which are much faster. For instance, the authors in \citep{a14030069} have developed a new time series modeling based on least squares FCM termed as LSFCM that is much faster rather than multi‐iteration stochastic searching. Accordingly, there is no exception for the proposed HFCM-FTS model in \cite{Orang2020} which has been trained by GA. Therefore, this part introduces novel methodology to overcome the mentioned fundamental problem. This model is composed of a group of randomized HFCM-FTS models termed as Random HFCM (or R-HFCM) model to predict univariate time series. The innovation of the proposed R-HFCM model is integrating the concepts of FCM and ESN reservoir computing method exploiting the least-squares regression algorithm as the learning strategy which is detailed in the following.

\begin{figure*}[!tbp]
    \centering
    \includegraphics[width=1\textwidth]{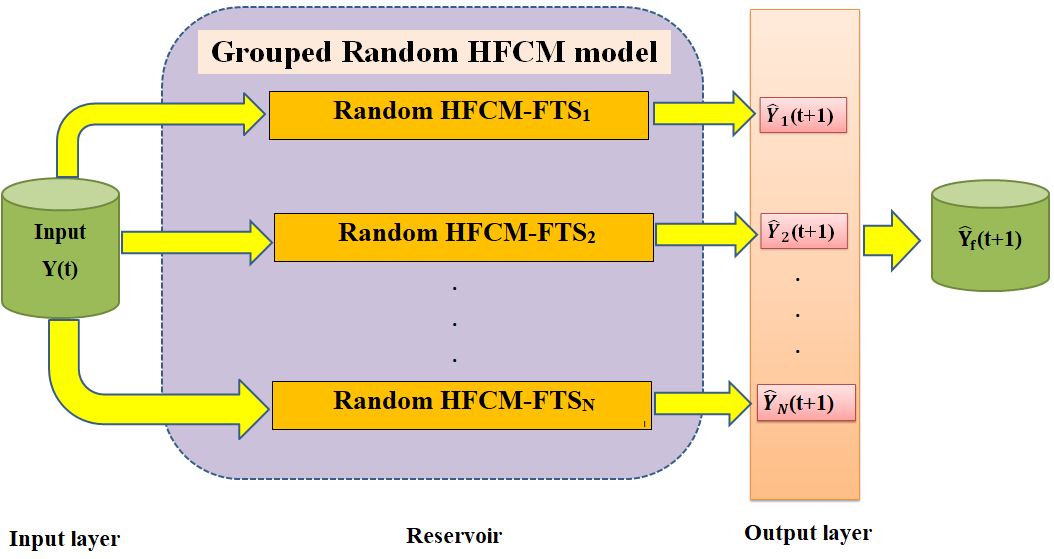}
    \caption{Generic structure of the proposed R-HFCM method.}
    \label{fig:GeneralRandomHfcm}
\end{figure*}

From a structural perspective, this model consists of three layers as illustrated in the Figure \ref{fig:GeneralRandomHfcm} including the input layer, the intermediate or reservoir layer and the output layer. The reservoir is composed of a group of random HFCM-FTS models that are fed only by the external input time series to provide inputs for the output layer. Then the obtained outputs from each sub-reservoir are considered as the units of the output layer that are used to generate the final predicted value. As deduced from Figure \ref{fig:GeneralRandomHfcm}, the reservoir structure consists of some numbers of sub-reservoirs ($N_{SR}$) so that there are no relations among them. It means that the proposed model does not have a hierarchical stacking structure similar to the deep models.

Since the structure of each sub-reservoir in the proposed R-HFCM model exactly is designed the same as the HFCM-FTS model,  the R-HFCM model is mainly concentrated on training the output layer using least squares regression. From another perspective, it can be said that the proposed R-HFCM is a kind of ESN in which only the output layer is trainable  while the reservoir parameters are initialized randomly and remain unalterable during the training process. On the other hand, the weights of each sub-reservoir in the reservoir layer are chosen randomly to meet the ESP condition in ESN and then each sub-reservoir generates its  output independently. It means that in our model the reservoir weights are initialized based on the ESP condition in ESN reservoir computing, then the least squares minimization algorithm is applied to the output units to generate the final predicted value. 

The proposed R-HFCM method is divided into two procedures: the Training procedure and the Forecasting procedure. Therefore, the Training and the Forecasting procedures are detailed in the rest of this section. 

\subsection{Training Procedure}

The main objective of the training procedure is to generate the linguistic variable $C$ with $k$ concepts, to find the best least square coefficients to fit the data,  given a crisp training set $Y$ and the activation function $f$ informed by the user. The steps of the method are listed below:

\begin{enumerate}
    \item \textbf{Partitioning}: 
    
    Since R-HFCM is composed of a group of HFCM-FTS models, the uniform scheme is applied for partitioning process in both R-HFCM and HFCM-FTS models but in this case for each sub-reservoir. More precisely, the Universe of Discourse (U), for each sub-reservoir, is partitioned into $k$ even length and overlapped intervals. Then, a fuzzy set $C_i$ (i.e. concept) is defined with a membership function $\mu_{C_i}$. In each sub-reservoir, the group of the $k$ concepts form the linguistic variable $C$, such that $C_i \in C$, $\forall i = 1,\ldots,k$. Here the grid partitioning is used to generate versions of the FCM with $k = \{5, 10, 20\}$, where the number of concepts is equal to the number of partitions and triangular membership function $\mu_{C_i}$. Figure \ref{fig:fcm} shows a simple example of the FCM structure used in this model. In Figure \ref{fig:fcm}-A the universe of discourse is partitioned to generate seven concepts, which is also the number of fuzzy sets because in our model the number of concepts and fuzzy sets is considered the same. As shown in Figure \ref{fig:fcm}-C, FCM is a collection of nodes (concepts) and causal interactions among these concepts which is represented via the weights matrix. According to Figure \ref{fig:fcm}-B, this weight matrix is a square connection matrix which is defined randomly through the following step.
    
    \begin{figure*}[!tbp]
        \centering
        \includegraphics[width=1\textwidth]{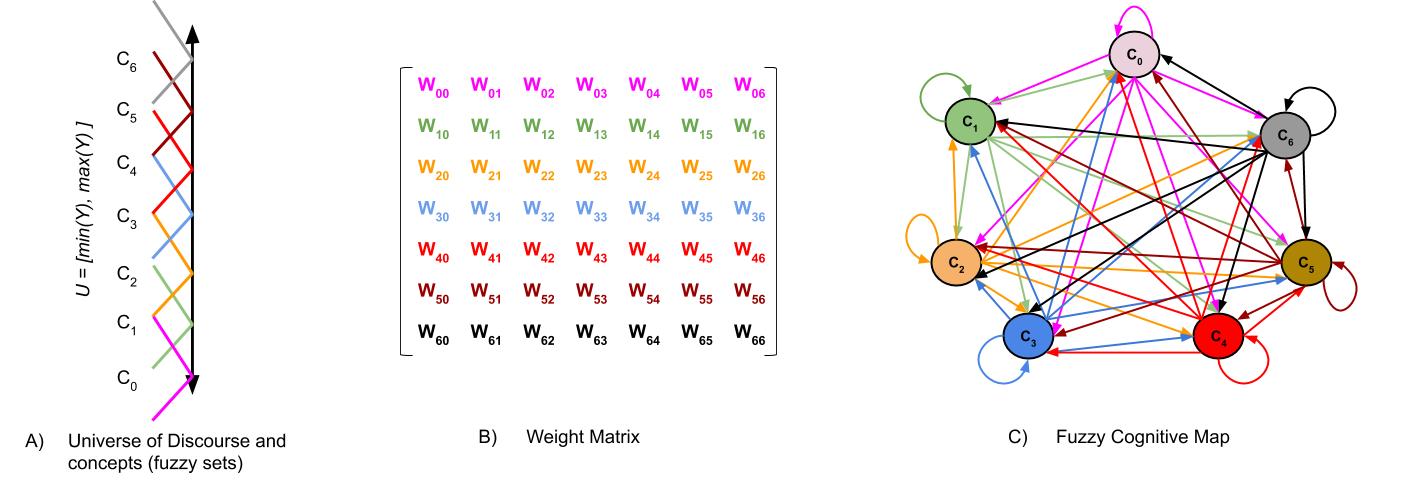}
        \caption{Structure of the fuzzy cognitive map}
        \label{fig:fcm}
    \end{figure*}
    
    \item \textbf{Least Squares coefficients determination}:
    
    Each matrix $\mathbf{W}^t$, for $t=1,\ldots,\Omega$, is a $k\times k$ matrix where $w_{ij}^t \in \mathbb{R}$ is the weight between the concepts $C_i$ and $C_j$ at the time lag $t$, and $\Omega$ is the order of the model. As noted earlier, the innovation in the proposed model is that the weight matrices of each HFCM-FTS are randomly chosen from a uniform distribution over interval [-1,1]. Then, they are scaled according to the ESN reservoir computing to preserve the ESP condition as described by the following formula: 
    \begin{equation}\label{eq:ESP}
    \mathbf{W}^t = \mathbf{W^{rand}}\cdot\left( \frac{\bf{\epsilon}}{\mathbf{\rho_{max}}(\mathbf{W^{rand})}} \right)
    \end{equation}
    where $\mathbf{\rho_{max}}(\mathbf{W^{rand})}$ is the maximum  eigenvalue of $\mathbf{W^{rand}}$ and $\mathbf{\epsilon} \in (0,1)$ is the desired spectral value (scaling parameter). Also, since the members of the weight matrix in the FCM must be in the range [-1,1], the value of $\mathbf{\epsilon} $ in our model is equal to $0.5$ to satisfy the corresponding condition. Hence, each reservoir generates its own output through the Forecasting procedure, given the training sample as well as the initialized weights. Finally, the least squares regression method is applied to train the last layer. Through this way, a linear model is solved and the optimum least-squares coefficients are obtained by minimizing the error function.
\end{enumerate}

\begin{figure*}[!tbp]
\centerline{\includegraphics[width=1.1\textwidth]{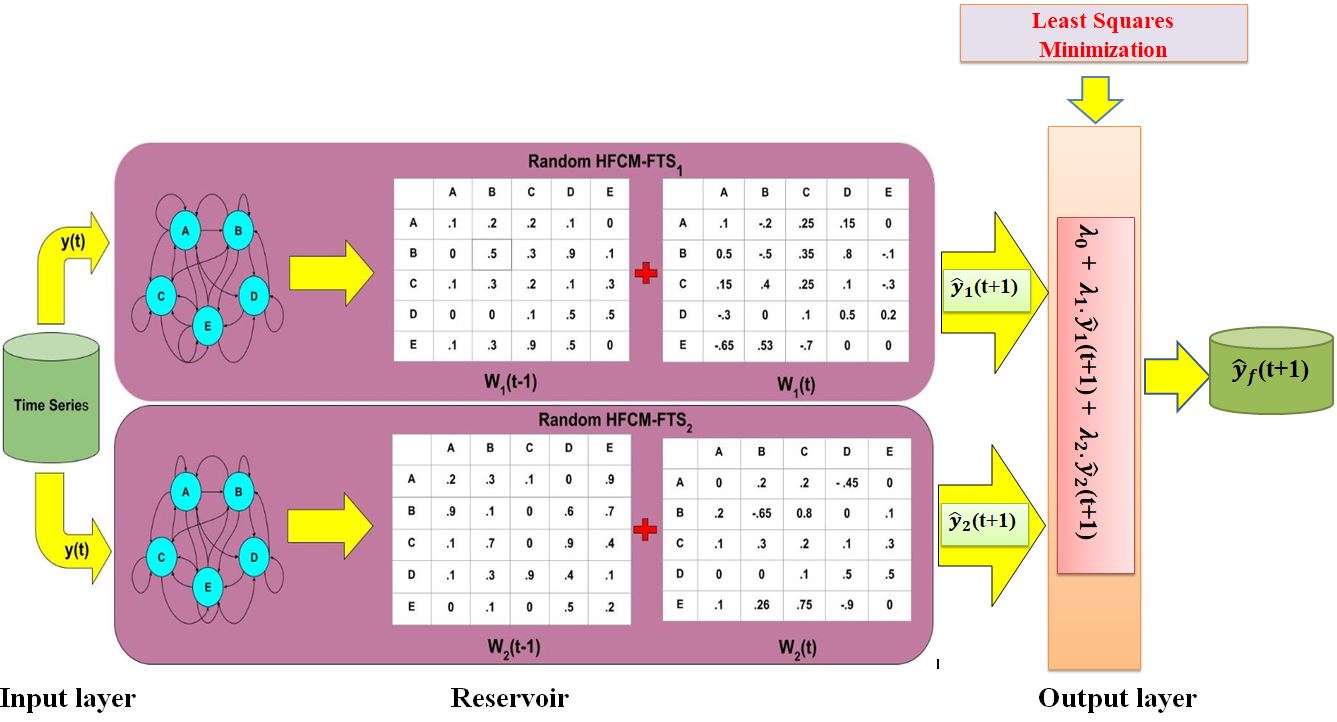}}
\caption{ The simple example of the proposed model mechanism considering $N_{SR}=2$. }
\label{fig:2HFCMs}
\end{figure*}

Figure \ref{fig:2HFCMs} shows a simple structure of R-HFCM model considering ${N_{SR}=2}$ while in each sub-reservoir the order is $\Omega = 2$. Thus there are two weight matrices for times $t$ and $t-1$ for each randomized HFCM-FTS as indicated in the figure. It is worth noting that the model contains the bias weights while they are discarded from the Figure \ref{fig:2HFCMs} just because of ease of notation.

\subsection{Forecasting Procedure}
\label{sec:forecasting_2}
The main goal of this stage is to compute the predicted crisp values $\hat{y}(t+1)$ of each sub-reservoir as well as the final predicted value $\hat{y}_{f}(t+1)$ by given the linguistic variable $C$, the weight matrices $\mathbf{W}^t$, the activation function $f$ and a crisp input $Y(t)$. The steps of this stage are listed below:

\begin{enumerate}
    \item \textbf{Fuzzification}: 
    
    In each sub-reservoir, given the crisp input sample $Y$ with size $T$, each instance $y(t) \in Y, t = 1..T$, is transformed into an activation vector $\mathbf{a}(t)$ such that $a_i(t) = { \mu_{C_i}( y(t) ) }$, $\forall C_i \in C$, that is, each value $a_i(t) \in \mathbf{a}(t)$ corresponds to the membership degree of $y(t)$ to the concept $C_i$.
    
    \item \textbf{Activation}:
    
    For each HFCM-FTS, the state value of each concept in time $t+1$ can be defined by the following formula:
    
    \begin{equation}\label{eq:two_order_fcm}
    \mathbf{a}(t+1) = f\left( \mathbf{w}^0+ \sum_{j=1}^\Omega \mathbf{W}^{j} \cdot \mathbf{a}(t-j+1)  \right)
    \end{equation}
    
    Since this model considers the presence of the bias term, $\mathbf{w}^0$ represents bias in the above equation. It should be noted that the initialization of the bias parameters and weight matrices are the same, such that the ESP condition in ESN reservoir computing is satisfied. It means that they are randomly chosen from a uniform distribution over the interval [-1,1], then rescaled according to the equation \eqref{eq:ESP} but in this case, the weight matrices are replaced by the bias ones.
    
    \item \textbf{Defuzzification}: 
    
    After calculating the activation level of each concept for each sub-reservoir, in this step, the defuzzification is carried out to produce the output related to each of them. Therefore, the forecast values for each sub-reservoir at time $t+1$ in numeric terms can be calculated via the below equation:
    \begin{equation}\label{eq:forecast_2}
    \hat{y}(t+1)=\dfrac{\sum_{i=1}^k a_{i}(t+1) \cdot mp_{i}}{\sum_{i=1}^k a_{i}(t+1)}
    \end{equation}
     where $a_i(t+1)$ is the activation calculated from the previous step for each concept at time $t+1$ and $mp_i$ is the center of each concept $C_i$.
     
     Finally, the predicted value at time $t+1$ is computed through the linear combination of the obtained outputs of all sub-reservoirs and the least squares coefficients. The final predicted value is described as following:
     
     \begin{equation}\label{eq:final_output}
    \hat{y}_{f}(t+1)=\lambda_0+\sum_{j=1}^{N_{SR}} \lambda_{j}\cdot \hat{y}_{j}(t+1)
    \end{equation}
    
     For instance, as highlighted in Figure \ref{fig:2HFCMs}, the final output is estimated via the below equation when $N_{SR}=2$ : 
     
    \begin{equation}\label{eq:final_output2}
    \hat{y}_{f}(t+1)=\lambda_0+\lambda_1 \cdot \hat{y}_{1}(t+1)+\lambda_2 \cdot \hat{y}_{2}(t+1)
    \end{equation}
    
    where $\{\lambda_0,\lambda_1,\lambda_2\}$ denote the obtained least square coefficients from training procedure and $\hat{y_1}(t+1)$ and $\hat{y_2}(t+1)$ are the generated outputs for the first and second sub-reservoirs respectively according to the equation \eqref{eq:forecast_2}. 
\end{enumerate}

As Figure \ref{fig:2HFCMs} recounts, the number of least squares coefficients directly depends on the number of sub-reservoirs $N_{SR}$. To be more precise, the number of least squares coefficients in this model is equal to $N_{SR}+1$. Thereby, for $N_{SR}=n$, the final predicted value is described by the following formula using $n+1$ least-squares coefficients.

\begin{equation}\label{eq:final_output3}
\hat{y_f}(t+1)=\lambda_0+\lambda_1 \cdot \hat{y_1}(t+1)+\lambda_2 \cdot \hat{y_2}(t+1)+...+\lambda_n \cdot \hat{y_n}(t+1)
\end{equation}

\section{Computational Experiments}
\label{sec:experiments}
\subsection{Dataset}

In order to evaluate the effectiveness of the proposed R-HFCM model, in this section, two different datasets including the SONDA dataset and the Malaysian dataset are employed.

\begin{table}[htb]
    \centering
    \begin{tabular}{|c|c|l|} \hline
        \textbf{Variable} & \textbf{Type} & \textbf{Description}  \\ \hline
        DateTime & Time Stamp & yyyy-MM-dd HH:MM  \\ \hline
        glo\_avg & Real & Global average solar radiation   \\ \hline
        ws\_10m & Real & Wind speed in meters by second (m/s)  \\ \hline
    \end{tabular}
    \caption{SONDA dataset variables}
    \label{tab:sonda_variables}
\end{table}
\begin{enumerate}[(A)]
\item \textbf{SONDA Dataset}: SONDA - \emph{Sistema de Organização Nacional de Dados Ambientais} (Brazilian National System of Environmental Data Organization), is a governmental project which groups environmental data (solar radiance, wind speed, precipitation, etc) from INPE - Instituto Nacional de Pesquisas Espaciais (Brazilian Institute of Space Research). This dataset was retrieved directly from the SONDA Project \footnote{\url{http://sonda.ccst.inpe.br/}}.

Since the R-HFCM model has been designed to predict univariate time series, to test the utility of the proposed method, we apply the model to predict solar radiation time series data\footnote{Available at \url{https://query.data.world/s/2bgegjggydd3venttp3zlosh3wpjqj} accessed on April 4th, 2020} (glo\_avg variable in the Table \ref{tab:sonda_variables}). Noteworthy that the minimum, maximum, average and standard deviation of the proposed solar radiance time series are $\{-6.0667, 1228.65,223.261,311.239\}$, respectively. 

\begin{figure}[!htbp]
\centerline{\includegraphics[width=1\textwidth]{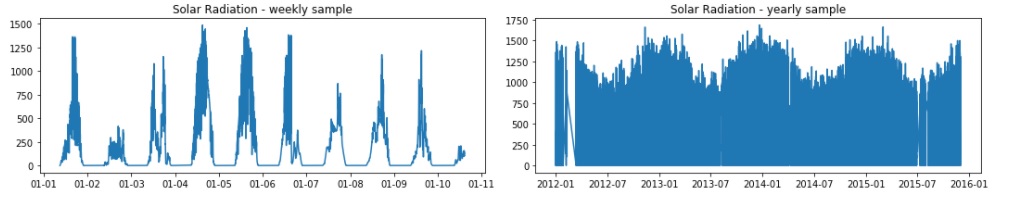}}
\caption{Weekly and yearly samples of solar radiance time series}
\label{fig:sonda_time_series_yearly_weekly}
\end{figure}

\begin{figure}[!htbp]
\centerline{\includegraphics[width=1\textwidth]{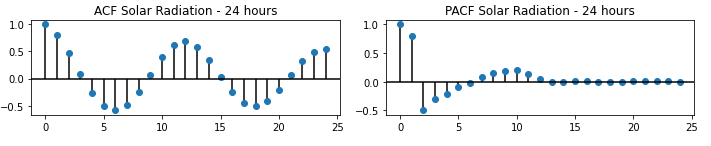}}
\caption{Autocorrelation and Partial Autocorrelation plots for solar radiance time series}
\label{fig:autocorrelation_time_series}
\end{figure}

\begin{figure}[!htbp]
\centerline{\includegraphics[width=0.7\textwidth]{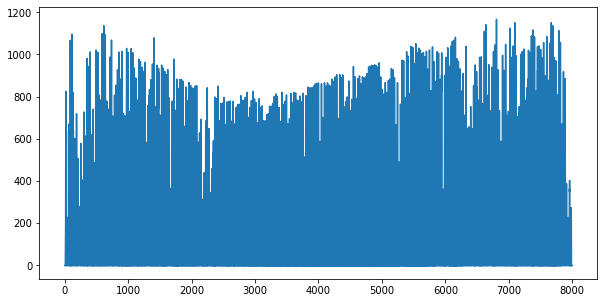}}
\caption{Samples of solar radiance time series}
\label{fig:time_series}
\end{figure}

\begin{table}[!htb]
    \centering
    \begin{tabular}{|c|c|l|} \hline
        \textbf{Variable} & \textbf{Type} & \textbf{Description}  \\ \hline
        DateTime & Time Stamp & yyyy-MM-dd HH:MM  \\ \hline
        temperature & Real & Temperature in Celsius degrees ($^o$C) \\ \hline
        load & Integer & Eletric load in Mega Watts by hour (MW/h)  \\ \hline
    \end{tabular}
    \caption{Malaysia dataset variables}
    \label{tab:malaysia_variables}
\end{table}

\begin{figure}[htbp]
\centerline{\includegraphics[width=1\textwidth]{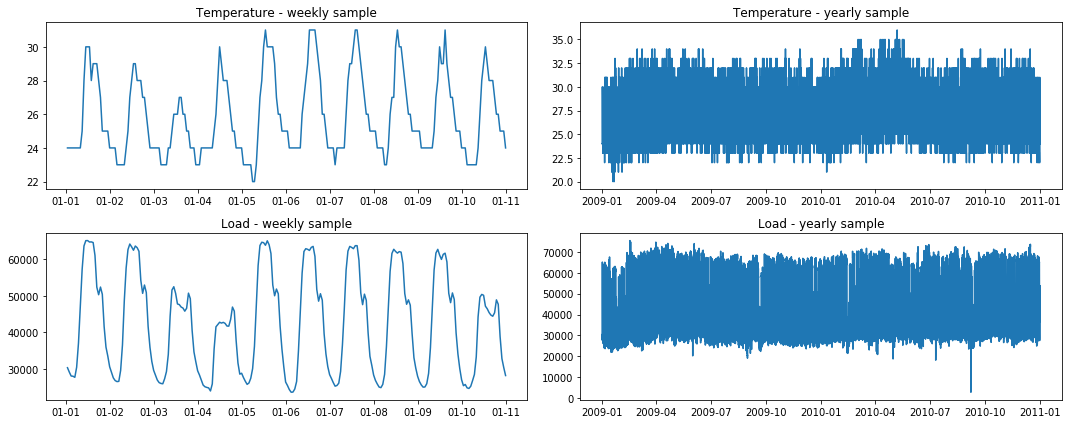}}
\caption{Weekly and yearly samples of Malaysia load and temperature time series}
\label{fig:samples_malaysia_yearly_weekly}
\end{figure}

\begin{figure}[htbp]
    \centering
    \includegraphics[width=\textwidth]{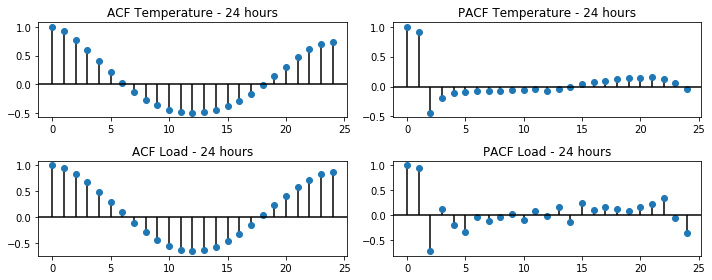}
    \caption{Autocorrelation and Partial Autocorrelation plots for Malaysia dataset}
    \label{fig:malaysia_acf}
\end{figure}

\begin{figure}[htbp]
    \centering
    \includegraphics[width=.7\textwidth]{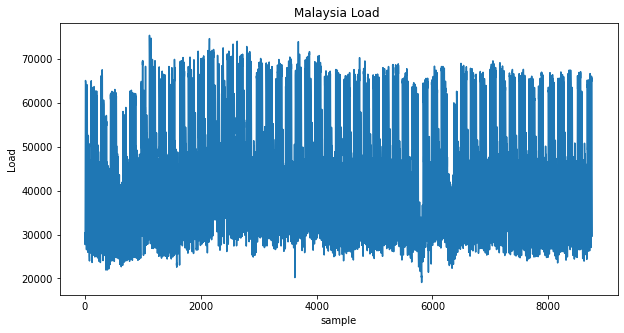}
    \caption{8760 Samples of Malaysia hourly load dataset }
    \label{fig:load}
\end{figure}

\begin{figure}[htbp]
\centerline{\includegraphics[width=.7\textwidth]{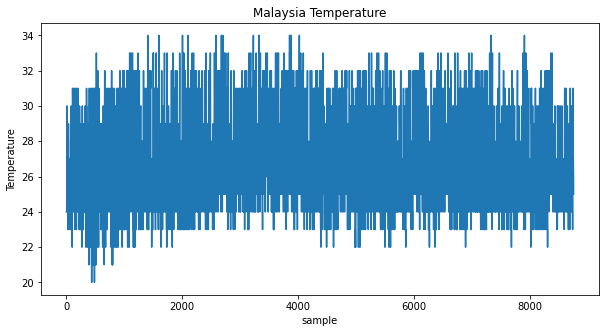}}
\caption{8760 Samples of Malaysia hourly temperature dataset}
\label{fig:temperature}
\end{figure}

Figure \ref{fig:sonda_time_series_yearly_weekly} shows the weekly and yearly samples of the referred solar radiance time series. Also, Figure \ref{fig:autocorrelation_time_series} highlights autocorrelation and partial autocorrelation plots for solar radiance time series. In this experiment, as shown in Figure \ref{fig:time_series}, 8,000 samples have been used, with a sliding window of 2,000 samples in the cross-validation method. 80\% of the window for training and 20\% for test.

\item \textbf{Malaysia Dataset}: As Table \ref{tab:malaysia_variables} exhibits, Malaysia dataset includes hourly electric load ($\{Min=2764,Max=75447, Ave=44323.167,std= 14212.389\}$) and temperature data ($\{Min=20,Max=36, Ave=26.606,std= 2.911\}$) of the power supply company of the city of Johor in Malaysia sampled between 2009 and 2010, with 17,519 instances which was retrieved from \cite{SADAEI2019365}. Figure \ref{fig:samples_malaysia_yearly_weekly} shows yearly and weekly samples of Malaysia dataset. Figures \ref{fig:load} and \ref{fig:temperature} show 8760 samples (year 2009) of the referred load and temperature time series.

\item \textbf{Evaluation Criteria}: For the quantitative evaluation of the proposed model quality, the performance indexes root mean squared error (RMSE), described in equation \eqref{eqn:rmse}, Mean Absolute Percentage Error (MAPE), described in equation \eqref{eqn:mape} and Theil’s U Statistic (U), described in equation \eqref{eqn:ustatistic} are used where $y(t)$ and $\hat{y}(t)$ stand for the actual and forecast values respectively.

\begin{equation}
RMSE = \sqrt{ \frac{1}{n} \sum_{i=1}^n (y_i - \hat{y}_i)^2 }
\label{eqn:rmse}
\end{equation}

\begin{equation}
MAPE = \frac{1}{n} \sum_{i=1}^n \Big|\frac{y_i - \hat{y}_i}{y_i}\Big|
\label{eqn:mape}
\end{equation}

\begin{equation}
U = \frac{\sqrt{\frac{1}{n}\sum_{i=1}^n (y_i - \hat{y}_i)^2}}{\sqrt{\frac{1}{n}\sum_{i=1}^n y_i^2} + \sqrt{\frac{1}{n}\sum_{i=1}^n \hat{y}_i^2}}
\label{eqn:ustatistic}
\end{equation}

\end{enumerate}

The performance of the proposed R-HFCM model is influenced by some factors. The number of concepts (partitions), the type of activation function, bias terms and the size of reservoir (the number of sub-reservoirs $N_{SR}$) are the significant parameters affecting the accuracy of the proposed method. Thus, in the following sections, we assess the performance of the proposed model for SONDA and Malaysia datasets respectively with respect to the mentioned influential parameters and compare the results with some recent state-of-the-art time series forecasting methods.

\begin{enumerate}[(A)]

\subsection{SONDA Case Study}
   \item \textbf{Results and Discussion}
   
In the first scenario, 8000 samples of the SONDA dataset have been used as can be seen from Figure \ref{fig:time_series}. A sliding window of $2,000$ samples in the cross validation method has been used. 80\% of the window for training and 20\% for test. The experiment has been executed by considering the presence of bias terms, $k=\{5,10,20\}$ and $N_{SR}=\{2,5,10,20,30,40\}$ to assess the model performance. Also, as can be seen from Table \ref{tab:fcm_activation}, different activation functions were used to design R-HFCM model.

\begin{table}[]
    \centering
\begin{tabular}{|c|c|} \hline
\textbf{Activation Function}    & \textbf{Mathematical Representation} \\ \hline

sigmoid  & 

$ f(x) = \frac{1}{\exp^{-{x}}+1}$ \\ \hline
hyperbolic tangent  & 
$ f(x) = \frac{\exp^{2x}-1}{\exp^{2x}+1} $\\ \hline

ReLU  & 
$f(x) = \max(0,x)$
    
     \\ \hline
softplus   & 
$ f(x) =  \ln({1+\exp^{x}}) $ \\ \hline

\end{tabular}
\caption{Proposed activation functions to design R-HFCM model}
    \label{tab:fcm_activation}
\end{table}

Table \ref{tab:SONDA DATA} provides information about  the average error in terms of RMSE and U for the 20 independent runs, given different activation functions, different number of concepts and sub-reservoirs. As the table shows, in all cases, the accuracy of the model is improved by increasing the number of sub-reservoirs for any number of concepts and regardless of the type of activation function. For instance, the model performs much better when $N_{SR}=40$ compared with $N_{SR}=5$. In return, increasing the number of concepts has an adverse effect on the accuracy of the model. It means that the more the number of fuzzy sets, the worse the model accuracy. For example, for every activation function and any number of $N_{SR}$, the model with 5 concepts performs more accurately  in contrast to the model with 20 concepts. In fact, as the results confirm, the accuracy of the model has been ameliorated by choosing the lower number of concepts as well as the higher number of sub-reservoirs. Since the weight and bias matrices are determined randomly on the basis of ESP in ESN, each configuration was tested 20 times.

To have a better understanding, Figures \ref{fig:SONDA_RMSE} and \ref{fig:SONDA_U} show a comparison of the model accuracy performance in terms of average RMSE and U respectively by taking the effective elements on the model accuracy into account including variant activation functions and both different number of concepts and sub-reservoirs. More precisely, each point represents the average value of RMSE or U after 20 experiments taking into account the predefined number of concepts and sub-reservoirs.

According to these figures, without considering the type of activation function, there is generally an inverse and direct relationship between the model accuracy with the number of concepts ($k$) and the number of randomized HFCM-FTS ($N_{SR}$) respectively. As is clear, with the constant value of $N_{SR}$, the minimum error is obtained with the least number of concepts. It is noticeable that the model performs slightly different  when $k=\{5,10\}$ and $N_{SR} \geq 20$ and also the error values (in terms of RMSE and U) do not become lower all along with the increasing of the map size.

\begin{center}
    \begin{landscape}
\begin{longtable}{|c|c|c|c|c|c|c|c|c|c|c|c|c|}
\hline
$\textbf{f} $& $\textbf{k}$ & $\textbf{N}_\textbf{SR} $ & 
 $\overset{\textbf{ Ave.}}{\textbf{ RMSE}} $ & $\overset{\textbf{ Ave.}}{\textbf{ U}} $ & $\textbf{k}$ & $\textbf{N}_\textbf{SR} $ & 
$\overset{\textbf{ Ave.}}{\textbf{ RMSE}} $ & $\overset{\textbf{ Ave.}}{\textbf{ U}} $& $\textbf{k}$ & $\textbf{N}_\textbf{SR} $ & 
$\overset{\textbf{ Ave.}}{\textbf{ RMSE}} $ & $\overset{\textbf{ Ave.}}{\textbf{ U}} $ \\ 
\hline

\multirow{6}{*}{sigmoid} & \multirow{6}{*}{5} & 2& 218.7964&  1.71 & \multirow{6}{*}{10} &  2 &  242.5481&  1.8899&
\multirow{6}{*}{20} & 2 &  258.0878& 2.0203
\\ 

& & 5& 159.6778& 1.237 &
& 5 & 200.9&  1.5657  &
& 5 & 224.3400&  1.7534  \\

& & 10&  104.005 &  0.7941 &
& 10& 158.9251& 1.2306 &
& 10& 195.2627& 1.5237\\

& & \textbf{20}& \textbf{98.8828}& \textbf{0.7533} &
& 20& 106.2003& 0.8107 &
& 20& 162.9512& 1.2648\\

& & 30 &101.4848& 0.7724 &
& \textbf{30}& \textbf{104.2803}& \textbf{0.7956} &
& 30& 142.752& 1.0838\\

& & 40 & 104.4596 & 0.7951& 
& 40& 109.9407&  0.8395 &
&\textbf{40}& \textbf{133.34}& \textbf{1.0161}\\

\hline
\multirow{6}{*}{ReLU} & \multirow{6}{*}{5} & 2 &226.0130& 1.7667  & \multirow{6}{*}{10} &  2&  251.6752&  1.9694 &
\multirow{6}{*}{20} & 2 &  261.0819& 2.040  
\\ 

& & 5& 173.6371& 1.3517  &
& 5 & 207.9632&  1.6197 &
& 5 & 229.5299 & 1.7924 \\

& & 10&  131.3070 & 1.0133  &
& 10& 173.272& 1.3474 &
& 10& 201.933 & 1.5722\\

& & 20 & 105.8692&  0.8097 &
& 20& 138.955& 1.0717 &
& 20& 175.1505& 1.3615\\

& & 30 & 100.4117& 0.7661 &
& 30& 120.8912& 0.9274 &
& 30& 156.644& 1.2140\\

& & \textbf{40} & \textbf{99.5396}& \textbf{0.7594} &
& \textbf{40}& \textbf{112.704}&  \textbf{0.8622} &
& \textbf{40}& \textbf{145.3438} &  \textbf{1.1231}\\
\hline
\multirow{6}{*}{Softplus} & \multirow{6}{*}{5} & 2 & 223.9737 & 1.7480 & \multirow{6}{*}{10} & 2 & 241.9253& 1.8887 &  
\multirow{6}{*}{20} & 2 &256.2766& 2.0038 
\\ 

& & 5& 159.5257 &1.2352 & 
& 5 & 198.3868 & 1.5447 &
& 5 & 222.9995&  1.741  \\

& & 10 &  104.0222&  0.7942 &
& 10&  159.4527& 1.2344 &
& 10& 194.6123& 1.5166\\

& & \textbf{20} & \textbf{98.5488}& \textbf{0.7506} &
& 20 & 106.4744& 0.8124 &
& 20& 161.7605&  1.2537\\

& & 30 & 100.6753& 0.7664 &
& \textbf{30} & \textbf{104.2682}& \textbf{0.7955} &
& 30& 133.573 &1.0203\\

& & 40 & 105.1047& 0.8353 &  
& 40 & 110.0458& 0.843 &
& \textbf{40}& \textbf{117.5980} &  \textbf{0.8974}\\

\hline
\multirow{6}{*}{tanh} & \multirow{6}{*}{5} &2 & 221.7707& 1.7327  & \multirow{6}{*}{10} & 2 &244.3244& 1.91 & 
\multirow{6}{*}{20} & 2 & 256.6119&  2.0069\\

& & 5 & 159.9527&  1.2422 &
& 5 &  200.1635& 1.5591 &
& 5 &  223.5872& 1.7454 \\

& & 10&  109.3229& 0.8369 &
& 10& 161.594&  1.2526 &
& 10& 192.7906& 1.5022 \\

& & 20 & 99.4457& 0.7576 &
& 20 & 111.2213 & 0.8499 &
& 20& 165.3205 & 1.2839 \\

& & \textbf{30} &\textbf{98.7783}&  \textbf{0.7522} &
& 30 & 104.8535&  0.8004 &
& 30&  138.7099 & 1.0657 \\ 

& & 40 &100.4214& 0.76432 &  
& \textbf{40} & \textbf{104.6511}& \textbf{0.7988} &
&\textbf{40} &  \textbf{120.0421}& \textbf{0.9179}\\ 
\hline
\caption{The model performance for SONDA time series  using \{5,10,20\} number of concepts, different HFCM numbers with respect to the different activation functions}
\label{tab:SONDA DATA}
\end{longtable}
\end{landscape}
\end{center}

\begin{table}[htbp]
\centering
\begin{tabular}{|c|c|c|c|c|c|}
\hline
$\textbf{f} $& $\textbf{k}$ & $\textbf{N}_\textbf{SR} $ & $\overset{\textbf{Best Ave.}}{\textbf{ RMSE}} $&$\overset{\textbf{Best Ave.}}{\textbf{ U}} $& $\overset{\textbf{ Num.}}{\textbf{ Parameters}} $ \\ 
\hline

\multirow{3}{*}{sigmoid} & 5 & 20 &98.8828 &0.7533& 21\\ 

& 10 & 30& 104.2803& 0.7956& 31\\

& 20& 40&133.44&1.0161&41\\

\hline

\multirow{3}{*}{ReLU} & 5 & 40& 99.5396& 0.7594& 41\\ 

& 10&40&112.704&0.8622&41\\

& 20&40&145.3438&1.1231&41\\

\hline
\multirow{3}{*}{\textbf{Softplus}} & \textbf{5}& \textbf{20}& \textbf{98.5488}& \textbf{0.7506}&\textbf{21} \\ 

& \textbf{10}& \textbf{30}&\textbf{104.2682}&\textbf{0.7955}&\textbf{31}\\

&\textbf{20}&\textbf{40}&\textbf{117.5980}&\textbf{0.8974}&\textbf{41}\\

\hline
\multirow{3}{*}{tanh} & 5& 30&98.7783&0.7522 &31\\ 

& 10&40& 104.6511& 0.7988&41\\

& 20& 40&120.0421&0.9179&41\\
\hline
\end{tabular}
\caption{The summary of the best performance of the model for SONDA dataset considering different f, while $k=\{5,10,20\}$ and $\Omega=2$ }
\label{tab:SONDA_summary}
\end{table}

\begin{figure}[htbp]
\centerline{\includegraphics[width=0.8\textwidth]{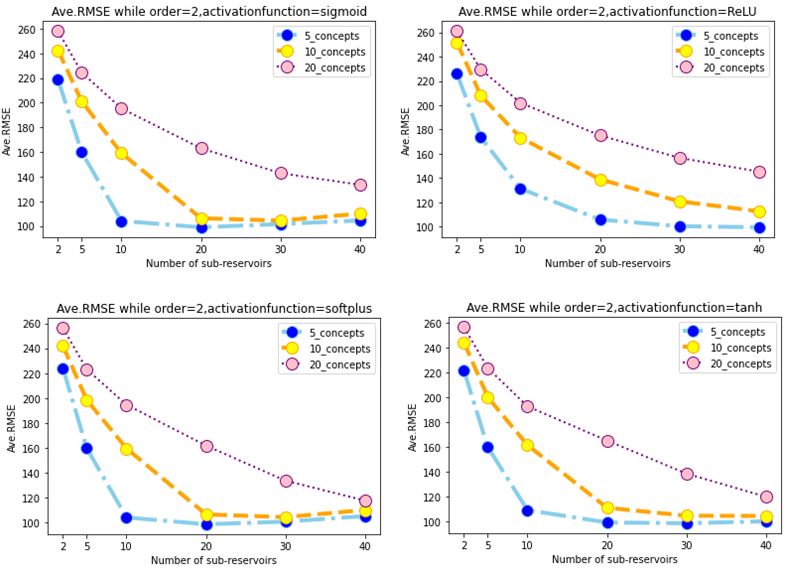}}
\caption{Comparison of the model performance in terms of the average RMSE considering different activation functions, different map sizes and different size of reservoirs after 20 independent experiments (SONDA data)}
\label{fig:SONDA_RMSE}
\end{figure}

On the other side, with a constant value of $k$, the error rate follows a downward trend as the number of $N_{SR}$ increases. Thus, the model is highly sensitive to the selection of both $k$ and $N_{SR}$ parameters and the optimal model is designed by increasing $N_{SR}$ and inserting $5$ concepts in each sub-reservoir. Also, as depicted in the figures, the effect of activation function over the model accuracy is less than the other factors. This means that there is no significant difference in the accuracy of the model when using different activation functions, especially in the case of using higher $N_{SR}$ and lower $k$. Therefore, as presented in Figure \ref{fig:SONDA_boxplot}, the model performance is less sensitive against the choice of the activation function. In this figure, each box contains six values in terms of average RMSE and U for different values of $N_{SR}$ with regards to the specified $k$ number and a certain type of activation function.

\begin{figure}[htbp]
\centerline{\includegraphics[width=0.8\textwidth]{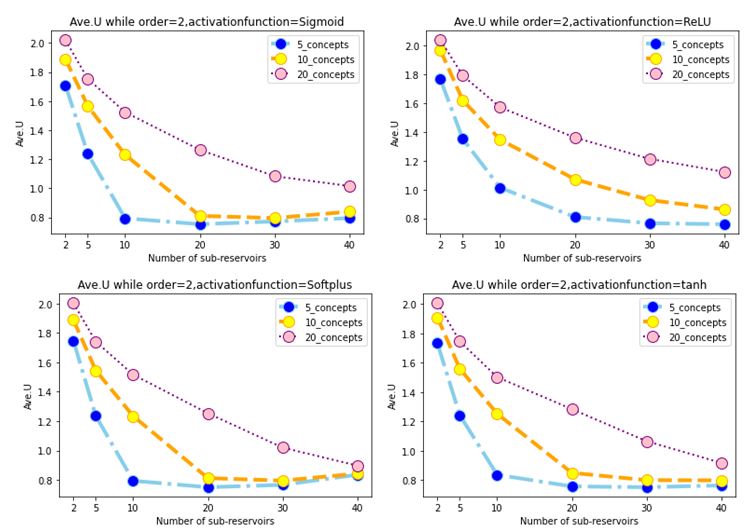}}
\caption{Comparison of the model performance in terms of the average U considering different activation functions, different map sizes and different size of reservoirs after 20 independent experiments (SONDA data)}
\label{fig:SONDA_U}
\end{figure}

\item{\textbf{Comparison with other Fuzzy Time Series methods}}

In this part, the proposed method is tested against other recent fuzzy time series forecasting methods available in the literature. In particular, the proposed model is tested against HFCM-FTS , HOFTS, WHOFTS and PWFTS methods. Bear in your mind that since the weights and bias are randomly chosen according to the ESN reservoir computing to meet the ESP condition, the R-HFCM results are obtained after 20 independent runs.  

Table \ref{tab:SONDA_summary} shows the data extracted from Table \ref{tab:SONDA DATA} on the basis of the best (minimum) average RMSE and U. Accordingly, Table \ref{tab:SONDA_summary} compares the best results obtained for each activation function with respect to the different number of concepts. More detailed look at the results  reveal that with the same map size $k$, the best performance can achieve with the same or different size of reservoirs. For example, when $ k =5 $ , the reservoir size of the models with ReLU and Softplus activation functions are respectively greater and equal to the size of the  reservoir of the model using sigmoid.

Although there exist slight differences among the model accuracy using different activation functions, irrespective of the number of concepts, the minimal error is achieved by considering Softplus as the model activation function. More precisely, selecting this activation function leads to the best model performance with unequal size of reservoirs such that for $k=\{5,10,20\}$ concepts, the optimum models are designed utilizing $N_{SR}=\{20,30,40\}$ sub-reservoirs, respectively. Therefore, the obtained results using Softplus are exploited to compare the suggested model with other recently presented methods in the literature. 

Accordingly, Table \ref{tab:accuracy2} compares the results of the proposed methodology  with the aforementioned methods with respect to the effective parameters. The results indicate the significant supremacy of the proposed R-HFCM over HFCM-FTS model. Actually, in comparison to the HFCM-FTS model, the proposed model is much faster and more accurate with fewer parameters. In other words, since the weights and bias matrices are determined randomly in the proposed model, the number of parameters only depends on the number of $N_{SR}$. Thus, the least squares regression algorithm in the proposed model will solve a linear problem with fewer variables compared with using GA in HFCM-FTS model. For instance, with 10 concepts, the HFCM-FTS parameters are more than 7 times that of the proposed R-HFCM model which can be considered as an effective factor on the training time of the model. Interestingly that by increasing the number of concepts from 10 to 20, only 10 parameters have been added to the proposed model, while this number will be almost quadrupled in the HFCM-FTS model. Additionally, adjustment of the learning parameters in GA is considered as another limitation compared with least squares method. Thereby, the training time of the model will be significantly affected by the mentioned factors.

\begin{figure}[htbp]
\centering
\includegraphics[width=1\textwidth]{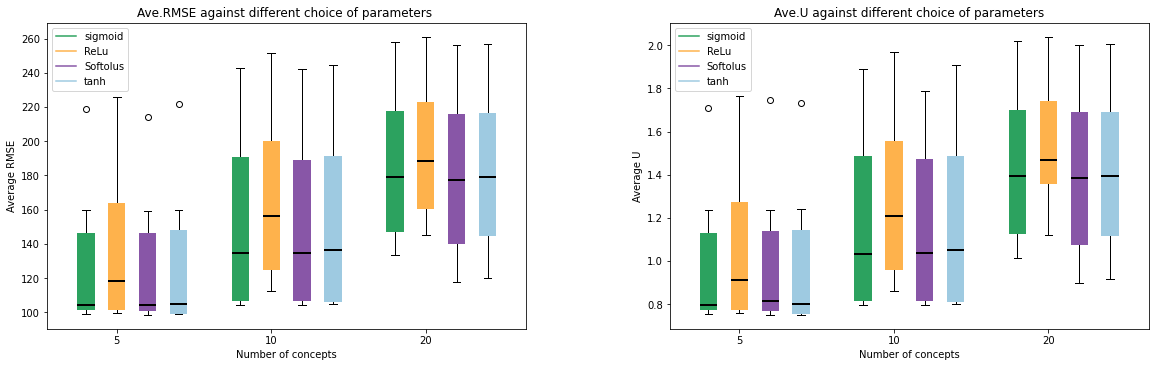}
\caption{Sensitivity of the proposed model accuracy against different choice of activation functions considering diverse values of  k and $N_{SR}$}
\label{fig:SONDA_boxplot}
\end{figure}

Unlike the HFCM-FTS model, which only outperforms other methods when $k = 5$, the R-HFCM model performs better than other methods regardless of the number of concepts. From the other viewpoint, similar to HFCM-FTS method, increasing the number of concepts degrades the accuracy of the model performance. Meanwhile, increasing the number of concepts meliorates the performance of HOFTS, WHOFTS and PWFTS models. 

\begin{table}[htbp]
\begin{center}
\resizebox{\columnwidth}{!}{\begin{tabular}{|c|c|c|c|c|c|c|c|}
\hline 
$\textbf{k} $ & $\overset{\textbf{R-HFCM}}{\textbf{Parameters}}$ & \textbf{R-HFCM} &  $\overset{\textbf{HFCM-FTS}}{\textbf{ Parameters}}$ & \textbf{HFCM-FTS} &  \textbf{HOFTS} & \textbf{WHOFTS} & \textbf{PWFTS} \\ 
\hline 
5 & \textbf{21} & \textbf{98.548} & 60 & 119.191 & 368.42 & 182.98
	 & 149.79\\ 
\hline 
10 & \textbf{31} & \textbf{104.2682} & 220 & 157.401 & 288.51 & 146.15 & 
	134.91 \\ 
\hline 
20 & \textbf{41} & \textbf{117.598} &	840 & 182.044 & 204.18 & 134.43 & 
134.22 \\ 
\hline 
\end{tabular}}
\caption{Evaluation of the accuracy of the  proposed  method against other models in terms of RMSE }

\label{tab:accuracy2}
\end{center}
\end{table}
\end{enumerate}

\subsection{Malaysia Case Study}\label{sec:Malaysia}

In the second scenario, in order to appraise the validity of the proposed model, we consider the Malaysia data set, which includes hourly electric load and temperature data. 8760 samples of load and temperature are selected as the model inputs as shown in Figures \ref{fig:load} and \ref{fig:temperature} respectively. Since the model has been designed to predict univariate time series, hourly electric load and temperature data are fed into the model separately. Thus, the rest of this section examines the obtained results using hourly electric load and temperature datasets, respectively.

\begin{enumerate}[(A)]
    \item {\textbf{Hourly electric load dataset}}
    
Similar to the first case study, a sliding window of $2,000$ samples in the cross validation method has been used but in this scenario 90\% of the window for training and 10\% for test. The experiment has been executed with $k=\{5,10,20\}$ concepts, $N_{SR}=\{2,5,10,20,30,40\}$ sub-reservoirs, Softplus activation function and by considering bias terms to assess the model performance. 

\begin{figure}[htbp]
\centerline{\includegraphics[width=0.9\textwidth]{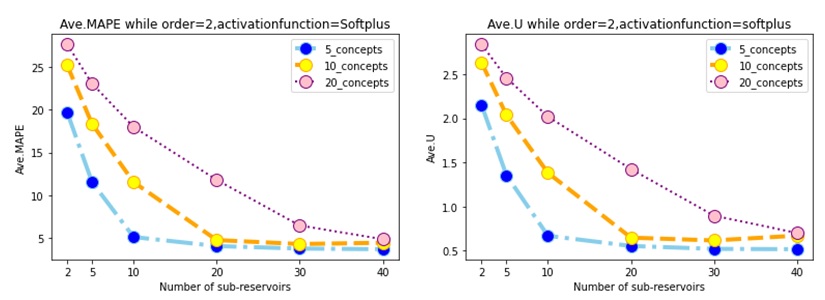}}
\caption{Comparison of the model accuracy in terms of (a) average MAPE and (b) average U by considering different $k$ and $N_{SR}$ when   activation function is Softplus (Malaysia hourly electric load dataset)}
\label{fig:load_MAPE_u}
\end{figure}

As illustrated in Table \ref{tab:Malaysia_load}, RMSE, MAPE and U are considered as forecasting accuracy metrics to assess the model performance. So it shows the average of each metric for the 20 independent runs given different numbers of concepts as well as reservoirs of different sizes. 

As the table summarizes, like SONDA case, the accuracy of the model has direct and indirect relation with $N_{SR}$ and $k$ respectively. The least value of the average error is generated with higher values of $N_{SR}$ for model with 5, 10 and 20 concepts, respectively. Also, Figure \ref{fig:load_MAPE_u} highlights clearly the relation between the model accuracy with $N_{SR}$ and $k$  in terms of the average MAPE and U by using Softplus activation function. More specifically, the error rate is decreasing from left to right. In simple terms, the higher value of $N_{SR}$, the more precise the forecasting model becomes. For instance, in all cases, the performance of the model is more accurate using $N_{SR}=40$ compared to the model with $N_{SR}=5$. Thus, with a fixed number of concepts, the higher the $N_{RS}$, the lower the forecasting error. In reverse, the lower the number of partitions, the better the model accuracy. It means that with a constant $N_{RS}$, increasing the number of $k$ leads to generate the model with less accuracy. Therefore, the lowest error rate is obtained with the minimum value of $ k $ and with increasing the size of the reservoir. Noteworthy, for $k=\{5,10\}$ and $N_{SR}\geq20$, the performance of the models are different slightly. For instance, for $k=5$ the model performances with $N_{SR}=\{30,40\}$ are very close.

\item{\textbf{Hourly temperature dataset}}

In this section, the codes are implemented to evaluate the model efficacy, while hourly temperature data is fed into the model as input instead of hourly load. In this case, like hourly load data, $ 2,000$ samples in the cross validation method have been used, with the same segmentation for training and testing. Also, this experiment is carried out with the same conditions, i.e.,  $k=\{5,10,20\}$, $N_{RS}=\{2,5,10,20,30,40\}$, Softplus activation function and the existence of bias terms.

\begin{figure}[htbp]
\centerline{\includegraphics[width=0.9\textwidth]{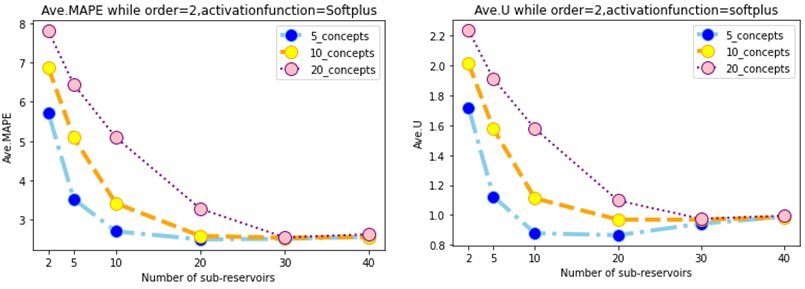}}
\caption{Comparison of the model accuracy in terms of (a) average MAPE and (b) average U by considering different $k$ and $N_{SR}$ when   activation function is Softplus (Malaysia hourly temperature dataset)}
\label{fig:temperature_MAPE_u}
\end{figure}

\begin{table}[htbp]
\centering
\begin{tabular}{|c|c|c|c|c|c|}
\hline
$\textbf{f} $& $\textbf{k}$ & $\textbf{N}_\textbf{SR} $ 
 & $\overset{\textbf{ Ave.}}{\textbf{ MAPE}} $ & $\overset{\textbf{ Ave.}}{\textbf{ RMSE}} $ &$\overset{\textbf{ Ave.}}{\textbf{ U}}$ \\ 
\hline

\multirow{6}{*}{Softplus} & \multirow{6}{*}{5} & 2 & 19.6603& 9849.6426& 2.1518 \\ 

& & 5&  11.5165 & 6204.0216 &1.3549 \\

& & 10& 5.1346 & 3069.7973& 0.67 \\

& & 20 & 4.0849 & 2540.4128& 0.555\\

& & 30& 3.7923 &  2390.6583 &  0.522  \\

& & \textbf{40} &\textbf{3.7189} &\textbf{2368.729}& \textbf{0.5181} \\
\hline
\multirow{6}{*}{Softplus} & \multirow{6}{*}{10} & 2 &25.1505& 12046.617&  2.6314 \\

& & 5&  18.3067& 9333.2505 &  2.0378 \\

& & 10&  11.5359&  6330.1563&  1.3826  \\

& & 20& 4.7811 & 2957.0319&  0.6489 \\

& & \textbf{30}& \textbf{4.3218}& \textbf{2806.5525}&  \textbf{0.6175}\\

& & 40& 4.4701& 3119.8898&  0.6916 \\

\hline
\multirow{6}{*}{Softplus} & \multirow{6}{*}{20} & 2 &  27.5919&  12991.5308& 2.8376 \\ 

& & 5&  23.0063 & 11254.5295&  2.4589 \\

& & 10 &17.9636& 9243.7015& 2.0192 \\

& & 20& 11.7987& 6523.8561 & 1.4255  \\

& & 30& 6.4828&  4094.8397&  0.8985 \\

& & \textbf{40}& \textbf{4.8811}&  \textbf{3180.6133}&  \textbf{0.7018} \\

\hline

\end{tabular}
\caption{The model performance for Malaysia hourly load data using different concept numbers and different number of layers when activation function is softplus}
\label{tab:Malaysia_load}
\end{table}
    
According to Table \ref{tab:Malaysia_temperature}, RMSE, MAPE and U are considered  as the accuracy metrics to evaluate the performance of the proposed method after 20 independent runs. The table records the  average  values of each accuracy metric applying the defined values of $k$ and $N_{SR}$.

As can be seen from  Table \ref{tab:Malaysia_temperature}, similar to the load case study, the relation between the model accuracy with $k$ and $N_{SR}$ follows the uniform pattern. That is, the fewer $k$ number and the more $N_{SR}$ number, the more accurate the model. Therefore, with a constant concept number $k$, the accuracy is improved by increasing the number of $N_{SR}$. For instance, assume that $k=5$, the model with $N_{SR}=30$ is much more accurate than the model with $N_{SR}=2$. On the other side, with a certain number of $N_{SR}$, boosting the number of concepts decays the model accuracy performance. For example assuming $N_{SR}=20$, the accuracy of the model is lessened by raising the value of $k$ from $5$ to $20$. 

These relations between the model accuracy with the effective elements on the model accuracy can also be deduced from Figure \ref{fig:temperature_MAPE_u}. It highlights that for any number of concepts the average errors are reduced by increasing the size of reservoir. Although the higher accuracy goes to the model with minimum $k$, the performance of the model with $k=\{5,10\}$ are very close when $N_{SR}\geq20$.

In addition, Figure \ref{fig:load_teperatre_boxplot} shows the box-plot that reflects the sensitivity of the method accuracy to the effective parameters including the map size and reservoir size for both load and temperature dataset. Each box consists of six values including the average error in terms of  MAPE or U considering $N_{SR}=\{2,5,10,20,30,40\}$ with respect to the defined number of concepts. As the figure recounts, increasing the number of $k$ deteriorates the accuracy. Overall, by considering the combination of the number of concepts and sub-reservoirs, it is clear that the best performance is achieved for the least values of $k$ and the higher number of $N_{SR}$. 

Accordingly, Table \ref{tab:best_results_load_temperature} summarizes the best model performance for both load and temperature data in terms of the best average errors. As the table confirms, by increasing the number of concepts, the model accuracy is declined. Also, for different number of concepts, the best results are obtained with the same or different size of reservoirs.

\item{\textbf{Comparison With Other Methods}}

In this section, the proposed R-HFCM model is tested against other investigated models in \cite{SADAEI2019365} under the assumption that the goal is to predict the hourly electric load. Accordingly, Table \ref{tab:comparison_2} shows the comparison among the obtained results from the proposed R-HFCM method and other models including FTS-CNN, LSTM, PWFTS, and SARIMA in terms of average  MAPE. R-HFCM is run with $k=5,10, 20$ concepts.

As the table suggested, the performance of the R-HFCM ($k=5$) and LSTM  are very close in terms of accuracy. It is interesting that, although the results indicate the superior performance of the FTS-CNN model over other methods, the results of the proposed R-HFCM model are still promising. This is because unlike FTS-CNN and LSTM methods, R-HFCM is a univariate model that utilizes only one variable (load) to predict hourly electric load. Also, R-HFCM model is a shallow model with lower parameters and cheap computational cost in comparison to the FTS-CNN and LSTM methods, which are deep learning models. Briefly, the proposed R-HFCM model which is equipped with a fast and non-repetitive learning algorithm is competitive with other FTS methods and some other state-of-the-art methods.  

In addition, in order to compare the performance of the proposed R-HFCM model, Figure \ref{fig:Friedman Aligned Ranks } represents the Friedman Aligned Ranks of the methods. The test statistic for these results is $Q=6.0$ where the p-value is 0.42319. For this statistic value, the $H_0$ is accepted at $\alpha=0.05$ confidence level which indicates that there is no difference between the means of the competitor models. In other words, all the methods are statistically equivalent. Therefore, the R-HFCM ($k=5$) method performed satisfactorily when compared with the standard methods in the literature.

\begin{table}[htbp]
\centering
\begin{tabular}{|c|c|c|c|c|c|}
\hline
$\textbf{f} $& $\textbf{k}$ & $\textbf{N}_\textbf{SR} $ 
 & $\overset{\textbf{ Ave.}}{\textbf{ MAPE}} $ & $\overset{\textbf{ Ave.}}{\textbf{ RMSE}} $ &$\overset{\textbf{ Ave.}}{\textbf{ U}}$ \\ 
\hline

\multirow{6}{*}{Softplus} & \multirow{6}{*}{5} & 2 & 5.7079 &  1.9635  &1.7182\\ 

& & 5& 3.5251&  1.2867&  1.1233\\

& & 10 & 2.7020 &  1.0062 &0.8778 \\

& & \textbf{20}&  \textbf{2.4993}  & \textbf{0.9917} & \textbf{0.8646} \\

& & {30} & 2.5121  & 1.0511 &  0.9419\\

& & 40 & 2.564&  1.0622& 0.987 \\

\hline
\multirow{6}{*}{Softplus} & \multirow{6}{*}{10} & 2 &6.8583&  2.3007&  2.0134 \\

& & 5& 5.0861 &  1.8019 &  1.5753  \\

& & 10  &3.4164 & 1.2748 &1.1115 \\

& & 20 &  2.5858 & 0.9960& 0.9693  \\

& & \textbf{30}&  \textbf{2.5371} &  \textbf{0.9956}&  \textbf{0.9686}\\

& &{40}& {2.5612} & {1.0216}&  {0.9853} \\

\hline
\multirow{6}{*}{Softplus} & \multirow{6}{*}{20} & 2 & 7.8018& 2.5549&  2.236 \\ 

& & 5&  6.4419& 2.1847&  1.9114  \\

& & 10& 5.0895& 1.8062& 1.578  \\

& & 20&  3.2790& 1.2624&  1.0965 \\

& & \textbf{30} & \textbf{2.5476} &\textbf{1.1068}&  \textbf{0.9745} \\

& & {40} & {2.6303}&  {1.1501}& {0.9951} \\

\hline
\end{tabular}
\caption{The model performance for Malaysia hourly temperature data using different map sizes and different number of layers while the activation function is softplus}
\label{tab:Malaysia_temperature}
\end{table}

\begin{figure}[htbp]
\centerline{\includegraphics[width=1\textwidth]{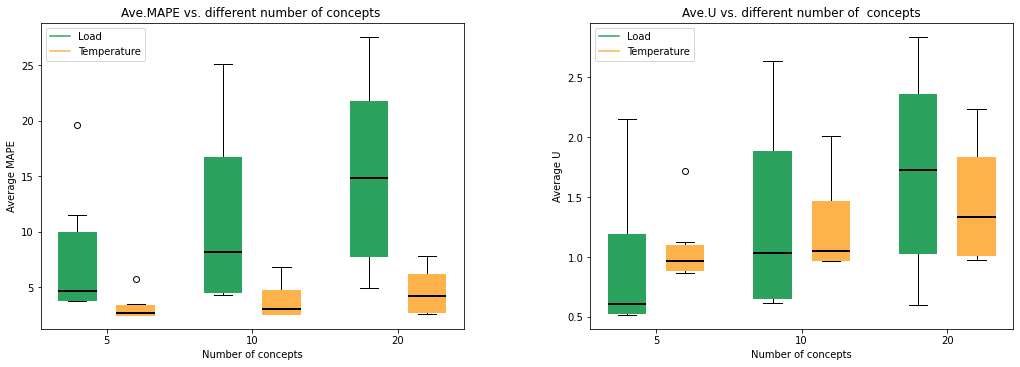}}
\caption{Comparison of the model accuracy in terms of average MAPE and U for both hourly load and temperature data considering effective parameters }
\label{fig:load_teperatre_boxplot}
\end{figure}
\begin{table}[htbp]
\centering
\begin{tabular}{|c|c|c|c|c|c|c|c|}
\hline
$\textbf{Dataset} $& $\textbf{f} $& $\textbf{k}$ & $\textbf{N}_\textbf{SR} $ &$\overset{\textbf{ Num.}}{\textbf{parameters}} $& $\overset{\textbf{ Best Ave.}}{\textbf{MAPE}} $& $\overset{\textbf{ Best Ave.}}{\textbf{RMSE}} $ & $\overset{\textbf{ Best Ave.}}{\textbf{U}} $ \\ 
\hline

\multirow{3}{*}{load} & \multirow{3}{*}{Softplus} & 5& 40 &41 &3.7189&2368.729&0.5181\\ 

& & 10& 30& 31 & 4.3218& 2806.5525& 0.6175 \\

& & 20&40&41& 4.8811&3180.6133&0.7018 \\

\hline

\multirow{3}{*}{temperature} & \multirow{3}{*}{Softplus} & 5 &20&21&2.4993&0.9917&0.8648  \\ 

& & 10&30&31&2.5371&0.9956&0.9686 \\

& & 20 & 30&31&2.5476&1.1068&0.9745\\

\hline
\end{tabular}
\caption{The performance of the model for both load and temperature Malaysia datasets in terms of the best average of the accuracy metrics for $k=\{5,10,20\}$}
\label{tab:best_results_load_temperature}
\end{table}

\begin{table}[]
    \centering
\begin{tabular}{|c|c|c|c|} \hline
\textbf{Methods}   & ${\textbf{ Uni/Multi}}{\textbf{ variate}} $ &   \textbf{Variable}&\textbf{Ave.MAPE}  \\ 
\hline
R-HFCM ($k=5$)&Univariate&load&3.7189\\ 
\hline
R-HFCM ($k=10$)&Univariate&load&4.1902 \\ 
\hline
R-HFCM ($k=20$)&Univariate&load&4.8377 \\ 
\hline
FTS-CNN &Multivariate& load+temperature& 3.02 \\ 
\hline
LSTM& Multivariate&load+temperature&3.71\\ 
\hline
PWFTS &Univariate& load&3.86\\ 
\hline
SARIMA &Univariate& load& 4.68 \\ 
\hline
\end{tabular}
\caption{Comparison of the proposed R-HFCM method with other models in term of average MAPE }
    \label{tab:comparison_2}
\end{table}

\begin{figure}[htbp]
\centerline{\includegraphics[width=.8\textwidth]{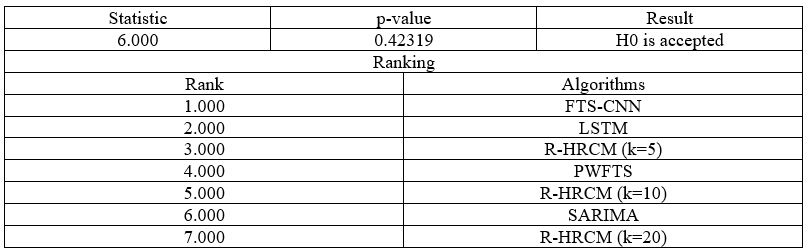}}
\caption{Friedman Aligned Ranks test (significance level of 0.05)}
\label{fig:Friedman Aligned Ranks }
\end{figure}

\end{enumerate}

\section{Conclusion and Future Works}
\label{sec:conclusion}

This paper provides a new randomized-based method to predict univariate time series using FCM. It is conducted based on the introduction of a novel FCM forecasting model in the literature termed as R-HFCM. 

R-HFCM model is a group of random HFCM-FTS models integrating the concepts of FCM and ESN. To be more clear, R-HFCM is composed of three layers employing random HFCM-FTS models as the components of the reservoir unit called sub-reservoirs. Since the structure of each sub-reservoir is exactly the same as HFCM-FTS, the main focus of this model is mainly on training the output layer using least squares regression. Thus, the weights of each sub-reservoir in the internal layer are initialized randomly so that the ESP condition in ESN is met and produces the output from each sub-reservoir. Then least squares regression method is applied to the output units to generate the final predicted value. Also, this method considers the effect of some elements on the model accuracy including the number of concepts, the number of sub-reservoirs, activation function and the presence of bias term. Based on the results, the more sub-reservoirs and the fewer concepts, the more accurate the model. That is the best result is achieved by increasing $N_{SR}$ and considering the minimum number of concepts in each sub-reservoir. Meanwhile, the activation function has the least effect on the accuracy of the proposed method compared with $N_{SR}$ and $k$.

It is considerable that R-HFCM offers a mixture of the advantages of high speed and accuracy over the HFCM-FTS model while the relation between the accuracy and the number of concepts is similar in both models. This means that as the number of concepts increases, the accuracy of the prediction, the interpretability and readability of the model deteriorates. Moreover, the number of parameters in R-HFCM directly depends on $N_{SR}$. Thereby, the least square method will solve a linear problem with fewer variables than GA, which can be counted as an effective element on the training speed of the model.

In the proposed R-HFCM model, the reservoir components are fed only through the external input to generate the input for the output layer. Then, the obtained results from each sub-reservoir are exploited to make the final predicted value. In this way, the proposed R-HFCM model has a shallow structure with no relations among the sub-reservoirs. The generated output from each layer of reservoir is only fed to the output layer of the model. Accordingly, the future challenge may involve extending the proposed model from shallow to deep one. Furthermore, as the model has been designed to predict univariate time series, predicting high-dimensional multivariate time series can be considered as the other possibility for future work. 

\section*{Acknowledgements}

\bibliography{mybibfile}

\end{document}